\documentclass{article}

\PassOptionsToPackage{numbers, compress}{natbib}
\usepackage[preprint]{neurips_2026}

\usepackage[utf8]{inputenc} 
\usepackage[T1]{fontenc}    
\usepackage{hyperref}       
\usepackage{url}            
\usepackage{booktabs}       
\usepackage{amsfonts}       
\usepackage{nicefrac}       
\usepackage{microtype}      
\usepackage{xcolor}         
\usepackage{multirow}
\usepackage[normalem]{ulem}
\usepackage[table]{xcolor}
\usepackage{graphicx}
\usepackage{amsmath}
\usepackage{diagbox}
\usepackage{wrapfig}
\usepackage{tabularx}
\usepackage{ragged2e}
\usepackage{makecell}
\usepackage{enumitem}
\usepackage[most]{tcolorbox}
\newtcolorbox{researchquestion}{
    colback=gray!4,
    colframe=black!60,
    boxrule=0.5pt,
    arc=1.5pt,
    left=4pt,
    right=4pt,
    top=4pt,
    bottom=4pt,
    before skip=5pt,
    after skip=5pt
}

\title{Different Prompts, Different Ranks: Prompt-aware Dynamic Rank Selection for SVD-based LLM Compression}

\author{%
Hengyi Zhu$^{1}$ \quad
Zhendong Mi$^{1}$ \quad
Grace Li Zhang$^{2}$ \quad
Shaoyi Huang$^{1}$\thanks{Corresponding author.} \\
$^{1}$Stevens Institute of Technology, Hoboken, USA \\
$^{2}$Technical University of Darmstadt, Darmstadt, Germany \\
\texttt{\{hzhu37, zmi2, shuang59\}@stevens.edu, grace.zhang@tu-darmstadt.de}
}

\begin{document}

\maketitle

\begin{abstract}
Large language models (LLMs) have rapidly grown in scale, creating substantial memory and computational costs that hinder efficient deployment. Singular value decomposition (SVD) has emerged as an effective post-training compression technique, but existing SVD-based methods rely on static rank truncation, applying a fixed prefix of singular components to all inputs regardless of their diversity. We identify two limitations of this static design: the optimal rank varies across individual prompts, and the selected rank is sensitive to the choice of calibration set, leading to suboptimal performance across diverse inputs. To address these challenges, we propose \textbf{PARSE}, a post-training framework for \textbf{P}rompt-\textbf{A}ware \textbf{R}ank \textbf{S}election as \textbf{E}xperts in SVD-compressed LLMs. PARSE trains a linear router offline to perform prompt-aware rank selection, decoupling it from calibration information by supervising the router against dense-model outputs on a large-scale corpus. We further observe that rank-selection patterns are shared across semantically similar prompts and remain stable across decoding steps, allowing appropriate rank subsets to be served directly from a pattern cache at inference. Complemented by expert memory aggregation and kernel fusion for system-level efficiency, PARSE is orthogonal to existing SVD-based pipelines and consistently improves both model quality and inference efficiency. Integrated with four representative SVD-based methods, PARSE improves average task accuracy by up to 10\% at a compression ratio of 0.6 on LLaMA-7B, and achieves up to 2.5 $\times$ prefill and 2.4 $\times$ decode speedup over native SVD execution.
\end{abstract}

\section{Introduction}

Large language models (LLMs) have become foundational across diverse applications, from natural language generation and translation to complex mathematical and code reasoning \cite{Guo_2025,hu2025openreasonerzeroopensourceapproach}. Guided by scaling laws \cite{kaplan2020scalinglawsneurallanguage}, modern LLMs such as GPT \cite{brown2020languagemodelsfewshotlearners}, PaLM \cite{chowdhery2022palmscalinglanguagemodeling}, LLaMA \cite{touvron2023llama}, DeepSeek \cite{Guo_2025}, and Qwen \cite{yang2025qwen3technicalreport} have grown to hundreds of billions of parameters. While such scale unlocks remarkable capabilities \cite{sheng2023flexgenhighthroughputgenerativeinference, ding2025dipsvd}, it also incurs heavy computational overhead that hinders inference on resource-constrained hardware and inflates the economic and environmental costs of large-scale serving \cite{zhou2024surveyefficientinferencelarge,wang2024modelcompressionefficientinference,fazlic2024enhancing,patterson2022carbon}. Improving LLM efficiency has therefore become a critical challenge for real-world deployment \cite{wang2025dobi,wang2024basis}. Common compression strategies, including quantization \cite{frantar2022gptq, lin2024awq, huang2024billm}, pruning \cite{ma2023llm, frantar2023sparsegpt, sun2023simple}, and distillation \cite{gu2024minillm, yang2025survey}, often depend on specific hardware \cite{dettmers2023qlora, lin2024awq} or require costly retraining \cite{bergmann2018polycystic, chen2020adabert}. In contrast, low-rank compression decomposes each weight matrix into three matrices and approximates it by discarding small singular values, directly reducing both parameter count and computational cost.

Recent advances in low-rank compression for LLMs, including ASVD \cite{yuan2023asvd}, SVD-LLM \cite{wang2024svd}, Basis Sharing \cite{wang2024basis}, and AdaSVD \cite{li2025adasvd}, have demonstrated that low-rank approximations can shrink model size with limited performance degradation. Nevertheless, existing approaches still face two notable challenges.  
\textbf{(1) Input-agnostic rank allocation.} Existing SVD-based methods typically determine the truncation rank through a one-shot calibration on a fixed dataset, yielding a single static rank applied uniformly to all inputs at inference time. This rigid allocation overlooks the fact that the rank required to faithfully preserve model behavior is itself input prompt-dependent, varying with the complexity and semantics of each prompt. A globally fixed rank is therefore inevitably suboptimal for some inputs and leads to unstable performance across heterogeneous prompts. \textbf{(2) Calibration-domain mismatch.} The static rank obtained from one-shot calibration is tightly coupled to the calibration set. When downstream tasks deviate semantically from this set, the chosen rank becomes misaligned with the target distribution, resulting in degraded performance on downstream inputs.


To address these challenges, we propose \textbf{PARSE}, a post-training framework for \textbf{P}rompt-\textbf{A}ware \textbf{R}ank \textbf{S}election as \textbf{E}xperts in SVD-compressed LLMs, orthogonal to existing SVD-based compression baselines. PARSE trains a linear router offline to learn prompt-aware rank selection and serves the selected rank subsets from a rank cache at inference. Specifically, we reformulate each SVD-compressed weight matrix as a mixture of rank experts, allowing the router to select a prompt-specific subset from a discrete set of independent rank components. To decouple rank selection from calibration information, the router is supervised against dense-model outputs on a large-scale corpus. To eliminate router overhead at inference, we retrieve cached subsets for semantically similar prompts and reuse the selected subset across decoding steps within the same prompt. We further introduce system-level optimizations, including expert memory aggregation and kernel fusion, to reduce runtime overhead.
%
%
Our contributions are summarized as follows:
\begin{itemize}[leftmargin=*, itemsep=2pt, topsep=0pt, parsep=1pt]

\item We identify two limitations of static rank truncation in SVD-compressed LLMs: (1) rank selection is sensitive to both individual input prompts within the same dataset and the choice of calibration dataset, leading to suboptimal performance.

\item We propose \textbf{PARSE}, a post-training framework that trains a linear router offline to perform prompt-aware rank selection, decoupling rank selection from calibration information by supervising the router against dense-model outputs on a large-scale corpus. \textbf{PARSE} is orthogonal to existing SVD-based compression baselines and consistently improves their performance.

\item We observe that rank patterns are shared across semantically similar prompts and remain stable across decoding steps within the same prompt, enabling rank subset retrieval at prefilling stage and rank reuse at decoding stage to reduce inference overhead. We further develop expert memory aggregation and kernel fusion to reduce system-level overhead.
\end{itemize}

\section{Preliminaries}
\label{sec:preliminaries}
For a weight matrix $W \in \mathbb{R}^{m \times n}$, its SVD is given by $W = U\Sigma V^\top$, where $U \in \mathbb{R}^{m \times m}$ and $V \in \mathbb{R}^{n \times n}$ are orthogonal matrices, and $\Sigma \in \mathbb{R}^{m \times n}$ is a diagonal matrix with singular values in descending order. Letting $r_{\max} = \min(m, n)$, this decomposes $W$ into a sum of rank components:
\begin{equation}
    W = \sum_{i=1}^{r_{\max}} \sigma_i u_i v_i^\top,
\end{equation}
where $\sigma_i$, $u_i$, and $v_i$ are the $i$-th singular value and left and right singular vectors. By the Eckart--Young--Mirsky theorem~\cite{eckart1936approximation}, retaining the top-$r$ components $\tilde{W} = \sum_{i=1}^{r} \sigma_i u_i v_i^\top$ minimizes the weight reconstruction error $\|\tilde{W} - W\|_F^2$ among all rank-$r$ approximations.

Minimizing weight reconstruction error does not, however, directly minimize activation reconstruction loss. The compression objective relevant to inference is:
\begin{equation}
    \|\tilde{W}X - WX\|_F^2,
\end{equation}
where $X \in \mathbb{R}^{n \times d}$ denotes the layer input activations over $d$ tokens.

To establish a direct correspondence between singular values and this loss, SVD-LLM~\cite{wang2024svd} introduces a data whitening step: a whitening matrix $S \in \mathbb{R}^{n \times n}$ is derived via Cholesky decomposition~\cite{meyer2023matrix} of $XX^\top$, enforcing $S^{-1}XX^\top(S^{-1})^\top = I$. SVD is then applied to $WS$ instead of $W$, under which the activation reconstruction loss from truncating the $i$-th component equals exactly $\sigma_i$. The compressed weight is recovered as $W' = U_r \Sigma_r V_r^\top S^{-1}$, and this pipeline has been widely adopted in subsequent work~\cite{wang2025dobi,hu2026saes}. In practice, $X$ is estimated from a static calibration dataset, and the rank-$r$ approximation is applied uniformly to all inputs at inference time.

\section{Observations}
\label{sec:obs}
Following SVD-LLM~\cite{wang2024svd} and subsequent works~\cite{wang2025svd,wang2025dobi,wang2024basis,hu2026saes}, existing SVD-based methods construct the whitening matrix from activations sampled from a calibration dataset. We identify two limitations of this static design.

\begin{wrapfigure}[14]{r}{0.4\textwidth}
    \centering
    \vspace{-0.25in}
    \includegraphics[width=0.4\textwidth]{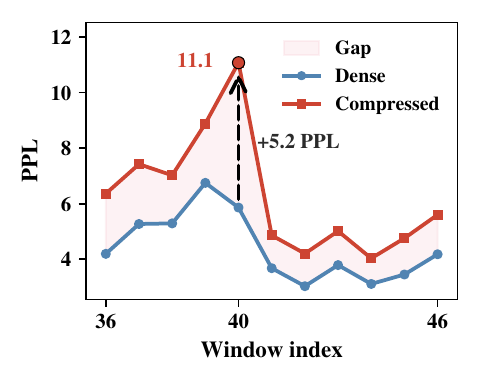}
    \vspace{-0.35in}
    \caption{
    Per-prompt window perplexity on WikiText-2 for the dense and compressed LLaMA-7B. 
    }
    \label{fig:ppl_spike}
    \vspace{-10pt}
\end{wrapfigure}

\textbf{Observation 1: Rank selection is sensitive to input prompts.}
Prior work~\cite{wang2025svd,wang2025dobi,wang2024basis,hu2026saes} typically evaluates compressed models by partitioning the test set into fixed-length windows and averaging perplexity (PPL) across them. Such aggregate metrics, however, mask the per-prompt behavior of the compressed model. Figure~\ref{fig:ppl_spike} reports the per-window PPL of the dense and compressed models on WikiText-2, where each window is a contiguous block of 2048 tokens. While the PPL increase is uniform across most windows, certain windows (e.g., index 40) exhibit sharp spikes that far exceed the rest, even though the dense model remains stable at these locations.
This indicates that static rank truncation discards components that are critical for specific prompts, causing localized degradation. The observation motivates our first research question (RQ1) as follows:

\begin{researchquestion}
\textbf{RQ1:} 
Can we design a rank selection strategy that dynamically adapts to each input prompt, retaining the critical singular components for that prompt?
\end{researchquestion}

\vspace{-0.4in}

\begin{figure*}[!ht]
    \centering
    \newcommand{\contentheight}{1.85in}

    \begin{minipage}[t]{0.34\linewidth}
        \centering
        \vbox to \contentheight{
            \vfil
            \includegraphics[width=\linewidth]{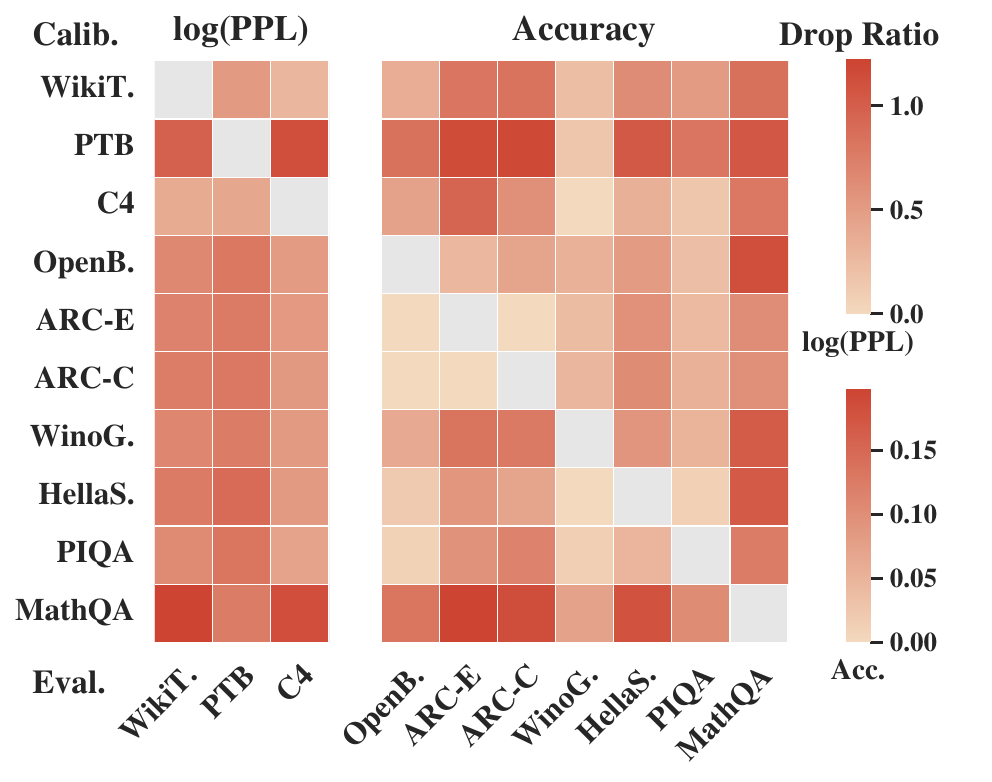}
            \vfil
        }
        \vspace{-0.15in}
        
            {\footnotesize (a)}
    \end{minipage}%
    \begin{minipage}[t]{0.24\linewidth}
        \centering
        \vbox to \contentheight{
            \vfil
            \scriptsize
            \setlength{\tabcolsep}{2.2pt}
            \renewcommand{\arraystretch}{0.92}
            \begin{tabular}{lrr}
                \toprule
                Calib. & $\Delta$PPL & $\Delta$Acc.(\%) \\
                \midrule
                WikiT.  & +5.30  & -9.81  \\
                PTB     & +18.81 & -14.48 \\
                C4      & +4.36  & -7.18  \\
                OpenB.  & +9.17  & -7.85  \\
                ARC-E   & +9.32  & -3.58  \\
                ARC-C   & +9.74  & -4.30  \\
                WinoG.  & +8.99  & -10.54 \\
                HellaS. & +10.58 & -6.04  \\
                PIQA    & +8.74  & -6.85  \\
                MathQA  & +18.35 & -14.49 \\
                \bottomrule
            \end{tabular}
            \vfil
        }
        \vspace{-0.15in}

        {\footnotesize (b)}
    \end{minipage}%
    \begin{minipage}[t]{0.35\linewidth}
        \centering
        \vbox to \contentheight{
            \vfil
            \includegraphics[width=\linewidth]{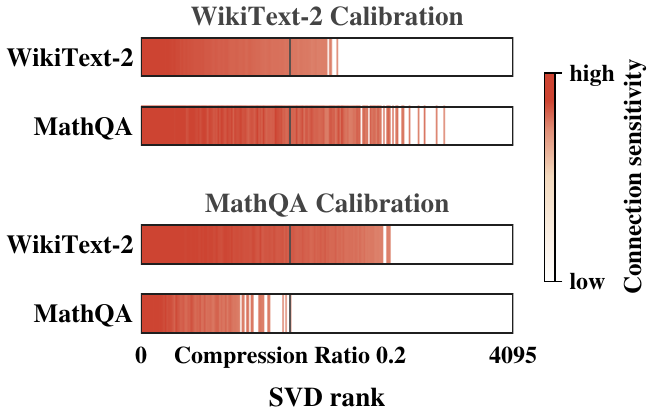}
            \vfil
        }
        \vspace{-0.15in}

        {\footnotesize (c)}
    \end{minipage}

    \vspace{-0.08in}
    \caption{
    (a) Cross-dataset evaluation of SVD-compressed LLaMA-7B under different calibration datasets.
    (b) Average off-diagonal degradation for each calibration dataset.
    (c) Gradient-based rank importance differs between WikiText-2 and MathQA for \textit{layers.19.self\_attn.o\_proj}, indicating that different calibration domains emphasize different singular components.
    }
    \label{fig:calibration_sensitivity}
    \vspace{-0.1in}
\end{figure*}

\textbf{Observation 2: Rank selection is sensitive to the choice of calibration dataset.}
We evaluate all combinations of calibration and evaluation datasets in Figure~\ref{fig:calibration_sensitivity}(a). Performance drops consistently when the evaluation dataset differs from the calibration set, with the largest gaps appearing when a specialized corpus such as MathQA or PTB is used for calibration, as reported in Figure~\ref{fig:calibration_sensitivity}(b). To diagnose the cause, we run the full SVD model without truncation and measure the rank-level connection sensitivity~\cite{lee2018snip} of each rank component across inputs from different tasks (Figure~\ref{fig:calibration_sensitivity}(c)). The top-ranked components remain consistently important across datasets, while the importance of tail ranks is highly dataset-dependent. For WikiText-2, the critical components concentrate in the top ranks, whereas MathQA relies on a different and more dispersed subset; the asymmetry holds in the reverse direction as well. At a compression ratio of 0.2, components essential for one dataset are truncated when calibrating on another. This indicates that the whitening matrix encodes the activation statistics of the calibration set, biasing the retained components toward calibration-specific information. The observation motivates our second research question (RQ2) as follows: 

\begin{researchquestion}
\textbf{RQ2:} 
Can we make rank selection independent of the choice of calibration dataset, such that the compressed model generalizes across diverse downstream datasets
?
\end{researchquestion}

\begin{figure}[t]
\centering
\includegraphics[width=\textwidth]{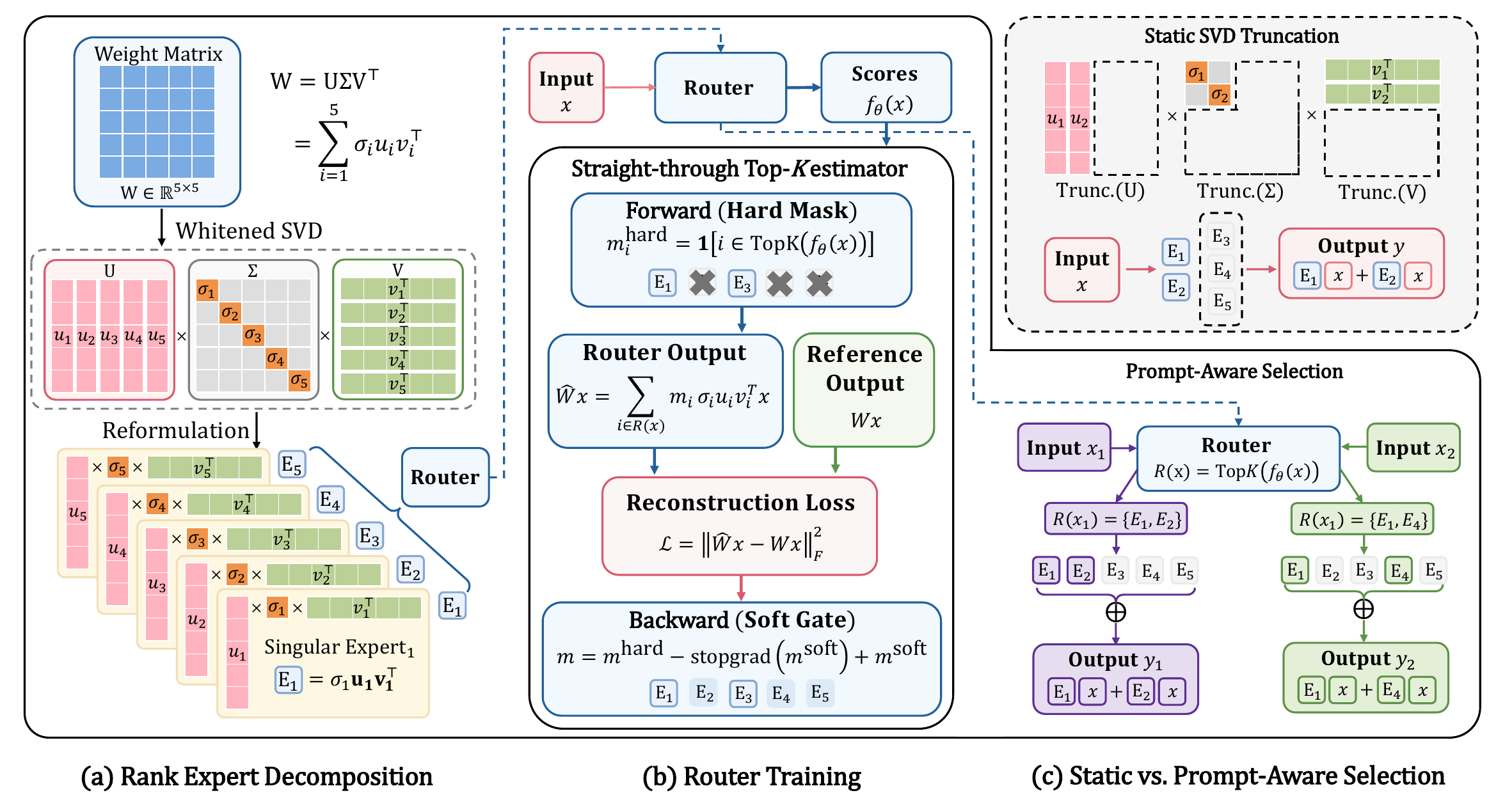}
\vspace{-0.2in}
\caption{Overview of the proposed framework with example of $W \in \mathbb{R}^{5 \times 5}$.
\textbf{(a) Rank Expert Decomposition.}
Each SVD-compressed weight matrix is reformulated as a set of independent rank experts.
\textbf{(b) Router Training.}
The router is optimized via a straight-through top-$K$ estimator with a reconstruction
loss against the dense model output.
\textbf{(c) Static vs. Prompt-Aware Selection.}
Static truncation retains a static subset for all inputs, while our router
selects a prompt-aware subset.
}
\label{fig:framework}
\vspace{-0.1in}
\end{figure}

\section{Method}
\vspace{-0.1in}
Motivated by these observations, we propose \textbf{PARSE}, a post-training framework that performs prompt-aware rank selection through an offline-trained router and cached rank retrieval at inference. We reformulate SVD-compressed weights as mixtures of rank experts, where the model dynamically retrieves prompt-aware rank subsets pre-optimized through offline router-based selection and truncation. 
The router is trained over the full-rank space of the dense model, enabling calibration-independent routing and allowing the model to select suitable rank subsets under different whitening and truncation settings. We further introduce rank retrieval and reuse strategies, memory aggregation and kernel fusion for efficient inference. Figure~\ref{fig:framework} provides an overview.



\begin{wrapfigure}[11]{r}{0.48\textwidth}
    \centering
    \vspace{-0.25in}
    \includegraphics[width=0.48\textwidth]{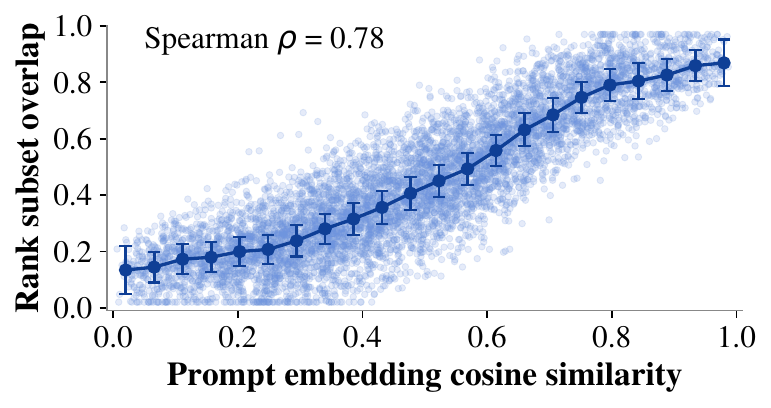}
    \vspace{-0.3in}
    \caption{Correlation between rank subset overlap and prompt embedding cosine similarity.}
    \label{fig:prompt_similarity_rank_overlap}
    \vspace{-0.25in}
\end{wrapfigure}

\vspace{-0.1in}
\subsection{Rank Expert Formulation for Prompt-Aware Selection}
\vspace{-0.05in}
\label{sec:router}
We reformulate each weight matrix as a mixture of independent rank experts and introduce an offline linear router that selects a prompt-aware subset for each input, addressing the static rank truncation in Observation 1 that discards components critical for specific prompts.
As established in Section~\ref{sec:preliminaries}, applying a weight matrix $W$ to an input activation $x$ can be written as a sum of rank contributions:
\begin{equation}
    Wx = \sum_{i=1}^{r_{\max}} \sigma_i u_i v_i^\top x,
\end{equation}
where $r_{\max}$ denotes the total number of rank components.
We treat each term $\sigma_i u_i v_i^\top$ as an independent rank expert $E_i$, as illustrated in Figure~\ref{fig:framework}(a).
Rather than statically retaining a fixed prefix, we select a prompt-aware index set $\mathcal{R}(x)$ of size $K$ from all rank experts. This defines a prompt-specific effective weight $\hat{W} = \sum_{i \in \mathcal{R}(x)} \sigma_i u_i v_i^\top$, whose output is:
\begin{equation}
\label{eqt:output}
    \hat{W}x = \sum_{i \in \mathcal{R}(x)} \sigma_i u_i v_i^\top x, \quad |\mathcal{R}(x)| = K,
\end{equation}
where $K < r_{\max}$ is the per-layer rank budget that controls the compression ratio.
To predict $\mathcal{R}(x)$, inspired by the gating mechanism in Mixture-of-Experts~\cite{fedus2022switch}, we attach a linear router $f_\theta: \mathbb{R}^{n} \to \mathbb{R}^{r_{\max}}$ to each weight matrix, where $n$ is the input dimension. The router produces a score $f_\theta(x) \in \mathbb{R}^{r_{\max}}$ over all rank experts given the input activation $x$, and $\mathcal{R}(x)$ is defined as the indices of the top-$K$ scores:
\begin{equation}
    \mathcal{R}(x) = \mathrm{TopK}(f_\theta(x)).
\end{equation}
As shown in Figure~\ref{fig:framework}(c), this allows different prompts to activate different rank subsets.
The per-layer budget $K$ is determined offline via Lagrangian multiplier-based optimization guided by each layer's information density~\cite{mi2025layer}, preserving the same global compression ratio as the baseline.

As shown in Figure~\ref{fig:prompt_similarity_rank_overlap}, prompts whose 
first-step prefill hidden states share high cosine similarity tend to produce 
highly overlapping router-selected rank subsets (Spearman $\rho = 0.78$), 
suggesting that rank selection patterns are largely determined by the semantic 
content of the prompt, already captured in the first-layer hidden representation. 
Consequently, at inference time we avoid invoking $f_\theta$ online: instead, 
we use the prefill hidden state as a prompt embedding to retrieve the nearest 
cached rank subset, as detailed in Section~\ref{sec:reuse}.

\subsection{Calibration-Decoupled Router Training}
\label{sec:training}
\paragraph{Dense-model-Supervised Calibration-Independent Router training.}
Following the post-training compression paradigm, we keep the SVD weights fixed and train only the router parameters $\theta$. 
Observation 2 shows that existing SVD-based methods perform a one-shot calibration on a specific dataset, which encodes dataset-specific activation statistics into the whitening matrix $S$. 
To decouple rank selection from calibration information, we train the router on a large-scale diverse corpus (e.g., C4) and supervise it against the output of the original dense model:
\begin{equation}
    \mathcal{L}_{\mathrm{rec}} = \left\| \hat{W}x - Wx \right\|_F^2,
\end{equation}
where $Wx$ is the reference dense output. Training on a diverse corpus encourages the router to learn generalizable rank-selection patterns across domains, while supervision from dense-model outputs aligns these patterns with the original model behavior rather than calibration-specific statistics. Consequently, regardless of the calibration dataset used for whitening and truncation, the router adaptively selects a suitable rank subset for each input, making rank selection substantially less sensitive to calibration data.

\vspace{-0.1in}
\paragraph{Straight-through Top-$K$ Estimator.}
Since the Top-$K$ selection is discrete and non-differentiable, gradients cannot flow back to the router parameters $\theta$ during training. We address this with a straight-through estimator that uses a hard mask in the forward pass and a differentiable surrogate in the backward pass. In the forward pass, we apply a binary mask $m_i^{\mathrm{hard}}$ to each rank expert:
\begin{equation}
    m_i^{\mathrm{hard}} = \mathbf{1}\!\left[i \in \mathcal{R}(x)\right],
\end{equation}
so that the prompt-aware output becomes $\hat{W}x = \sum_{i} m_i^{\mathrm{hard}}\, \sigma_i u_i v_i^\top x$, which is equivalent to Equation~\eqref{eqt:output}. In the backward pass, the effective mask $m \in \mathbb{R}^{r_{\max}}$ used for gradient computation substitutes $m^{\mathrm{hard}}$ with a differentiable surrogate $m^{\mathrm{soft}}$:
\begin{equation}
    m = m^{\mathrm{hard}} - \mathrm{stopgrad}(m^{\mathrm{soft}}) + m^{\mathrm{soft}}.
\end{equation}
Here $\mathrm{stopgrad}(\cdot)$ returns its input value in the forward pass but blocks gradient propagation in the backward pass. We define $m^{\mathrm{soft}}$ via sigmoid gating over the full set of rank experts:
\begin{equation}
    m_i^{\mathrm{soft}} = K \cdot
    \frac{\mathrm{sigmoid}(f_\theta(x)_i / \tau)}
         {\sum_{j} \mathrm{sigmoid}(f_\theta(x)_j / \tau) + \epsilon},
\end{equation}
where $\tau$ is a temperature hyperparameter and $\epsilon$ ensures numerical stability. The $K$-normalization keeps the expected gate magnitude consistent with the hard mask, while full-set coverage ensures all experts receive gradient signal regardless of whether they are currently selected.

\subsection{Rank Retrieval for Prefilling and Reuse for Decoding}
\label{sec:reuse}

While the router enables prompt-aware rank selection, it introduces overhead at both prefilling and decoding stages. We identify one property for each stage that allows this cost to be avoided, and validate each empirically.

\paragraph{Rank Retrieval for Prefilling.}
At each prefilling stage, invoking the router incurs an additional forward pass through $f_\theta$ for every incoming prompt. We therefore investigate whether rank patterns can be shared across prompts, avoiding per-prompt routing entirely. Figure~\ref{fig:prompt_similarity_rank_overlap} shows that semantically similar prompts tend to produce highly similar router-selected rank subsets. Motivated by this, we construct a rank pattern cache that maps representative prompt embeddings to their corresponding rank subsets. At inference time, the incoming prompt is matched to its nearest neighbor in the cache, and the associated rank subset is retrieved directly without invoking the router.

\begin{wrapfigure}[10]{r}{0.45\textwidth}
    \centering
    \vspace{-0.2in}
    \includegraphics[width=0.45\textwidth]{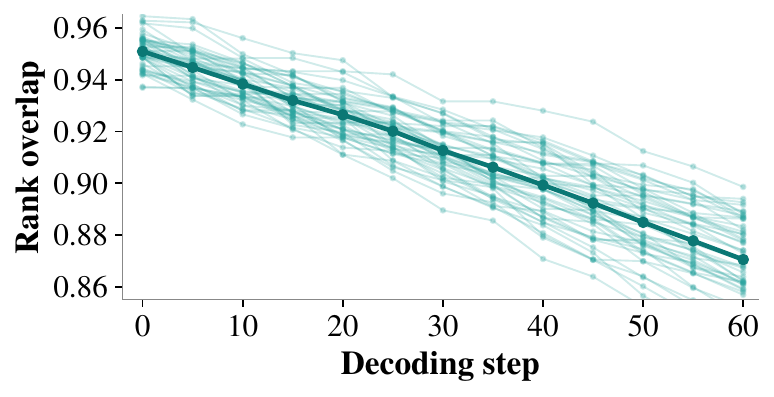}
    \vspace{-0.3in}
    \caption{Rank overlap between the subset selected at prefilling and those selected at each subsequent decoding step.}
    \label{fig:rank_stability}
    \vspace{-0.25in}
\end{wrapfigure}
\vspace{-0.1in}
\paragraph{Rank Reuse for Decoding.}
At each decoding step, querying the router introduces an additional forward pass through $f_\theta$ before every matrix computation, accumulating latency overhead across all layers and steps. As shown in Figure~\ref{fig:rank_stability}, the rank subset selected at prefilling remains highly consistent throughout generation, with rank overlap staying above $0.86$ across 60 decoding steps. This allows us to fix the rank subset at prefilling and reuse it for all subsequent decoding steps, eliminating router overhead entirely during generation.


\vspace{-0.1in}
\subsection{Memory Aggregation and Kernel Fusion}
\vspace{-0.1in}
\label{sec:system}
Following the pipeline in Section~\ref{sec:preliminaries}, we absorb $\Sigma$ into the left singular matrix and $S^{-1}$ into the right singular matrix offline:
\begin{equation}
\small
A=U\Sigma\in\mathbb{R}^{m\times r},\quad
B=S^{-\top}V\in\mathbb{R}^{n\times r},\quad
Wx=A(B^\top x)=\sum_{i=1}^{r}a_i b_i^\top x .
\end{equation}
Here, $r$ is the per-layer rank budget, and $a_i\in\mathbb{R}^{m}$ and $b_i\in\mathbb{R}^{n}$ are the $i$-th columns of $A$ and $B$, respectively. This removes $\Sigma$ and $S^{-1}$ from online computation, leaving two sequential MatMuls; each $a_i b_i^\top$ corresponds to one rank expert $E_i$ defined in Section~\ref{sec:router}.

\begin{wrapfigure}[21]{r}{0.6\textwidth}
    \centering
    \vspace{-0.3in}
    \includegraphics[width=0.6\textwidth]{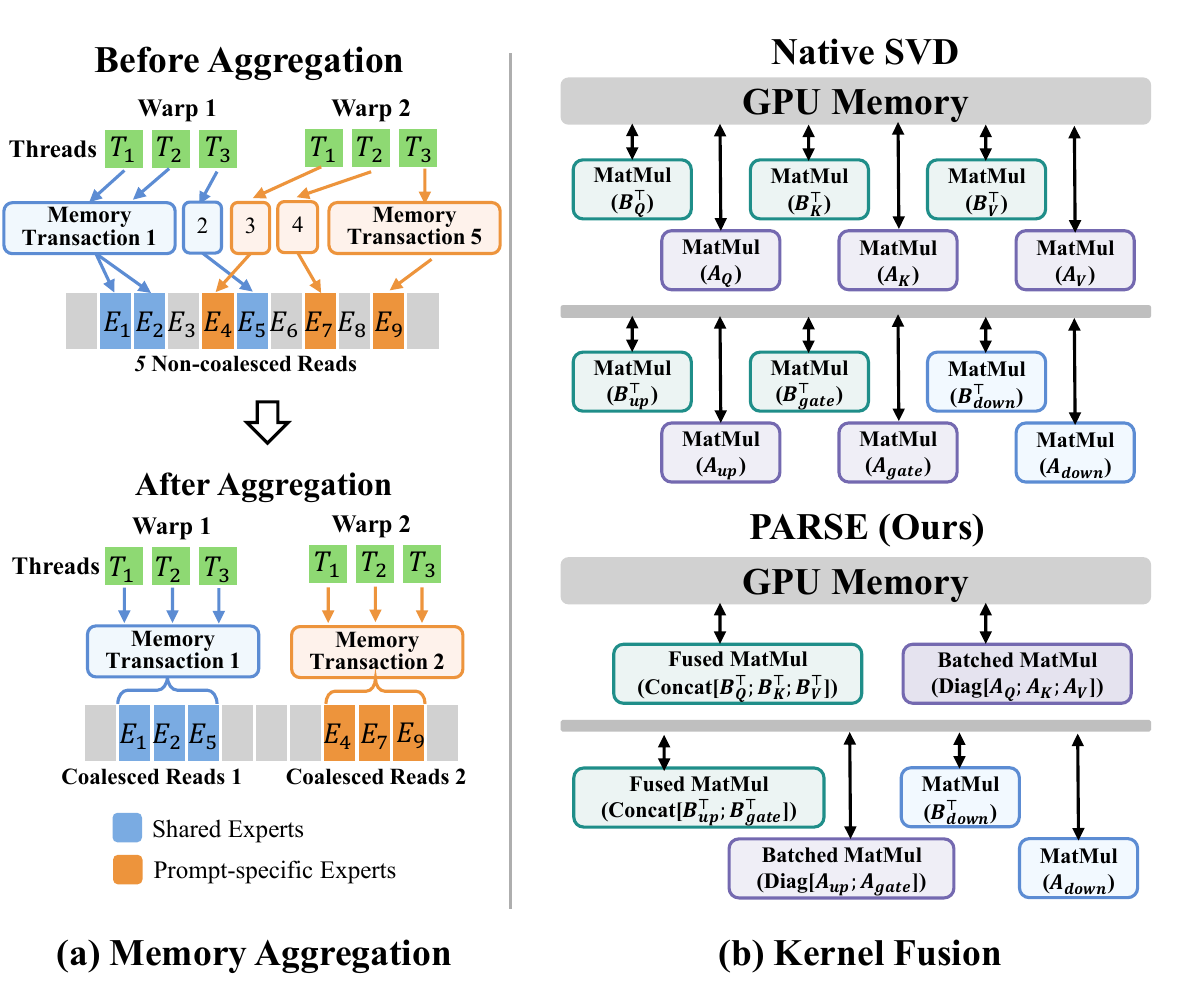}
    \vspace{-0.3in}
    \caption{(a) Memory aggregation reduces scattered expert reads to coalesced transactions. (b) Kernel fusion reduces redundant MatMul launches.}
    \label{fig:system_optimization}
    \vspace{-0.25in}
\end{wrapfigure}

\vspace{-0.1in}
\paragraph{Expert Memory Aggregation.}

Since the rank components selected by $\mathcal{R}(x)$ are physically scattered across $A$ and $B$, they fall on different cache lines, preventing GPU memory coalescing and inflating memory transactions. To address this, we reorganize the rank components into contiguous memory blocks once offline, as shown in Figure~\ref{fig:system_optimization}(a). Components with the largest singular values are selected by nearly all prompts; we designate these as \textit{shared experts} and place them at the head of memory, followed contiguously by the remaining prompt-specific components. As a result, $K$ scattered reads collapse into two coalesced memory accesses, with no change to the forward computation.

\vspace{-0.1in}
\paragraph{Kernel Fusion.}
SVD compression decomposes each weight matrix into two MatMuls per projection, doubling the number of kernel launches relative to the original dense layer. Since this overhead accumulates across all projections and layers, we reduce it by fusing operations that share the same input as shown in Figure~\ref{fig:system_optimization}(b).

\textbf{$B^\top$-side.}
The $B^\top$-side projections of $Q$, $K$, $V$ in the attention module share the same input activation $x$, as do the up- and gate-projections in the MLP module. We stack their $B^\top$ matrices and issue one fused MatMul per module, producing low-rank intermediates $z_i = B_i^\top x$:
\begin{equation}
\small
[[z_Q;z_K;z_V]]
=
[(B_Q)^\top;(B_K)^\top;(B_V)^\top]x,
\qquad
[z_{\mathrm{up}};z_{\mathrm{gate}}]
=
[(B_{\mathrm{up}})^\top;(B_{\mathrm{gate}})^\top]x .
\end{equation}

\textbf{$A$-side.}
Each intermediate $z_i$ is then projected by its corresponding $A_i$. Since these projections are independent across projections, they are batched into a single batched MatMul kernel per module:
\begin{equation}
\small
\hspace*{-0.01\linewidth}
[Q;K;V]=\operatorname{Diag}(A_Q,A_K,A_V)[z_Q;z_K;z_V],
\quad
[\mathrm{up};\mathrm{gate}]
=\operatorname{Diag}(A_{\rm up},A_{\rm gate})[z_{\rm up};z_{\rm gate}].
\end{equation}
For models with grouped query attention (GQA)~\cite{ainslie2023gqa}, where $K$ and $V$ have fewer heads than $Q$, the $A$-side of $Q$ cannot be batched with $K$ and $V$ due to mismatched output dimensions, increasing the attention-side batched MatMul count from 1 to 2.

\section{Experiment}
\subsection{Experimental Setup}
\label{sec:exp_setup}
\textbf{Models and Datasets.} We conduct experiments on open-source LLMs and widely used language modeling and zero-shot reasoning benchmarks. For model selection, we consider representative architectures including LLaMA-7B, LLaMA-13B, LLaMA-30B \cite{touvron2023llama} and Qwen2.5-7B \cite{hui2024qwen2}. For evaluation, we report perplexity on WikiText2 \cite{merity2016pointer}, PTB \cite{marcus1993building}, and C4 \cite{raffel2020exploring}, and zero-shot accuracy on OpenBookQA \cite{mihaylov2018can}, ARC-e, ARC-c \cite{clark2018think}, WinoGrande \cite{sakaguchi2021winogrande}, HellaSwag \cite{zellers2019hellaswag}, PIQA \cite{bisk2020piqa}, and MathQA \cite{amini2019mathqa}. All downstream reasoning tasks are evaluated in the zero-shot setting using the LM-Evaluation-Harness \cite{eval-harness}.

\vspace{-0.05in}
\textbf{Baselines.} Our method is not a standalone matrix decomposition framework, but a rank selection mechanism built on top of existing SVD-based compression methods. Therefore, we evaluate it by integrating it with several representative and influential SVD-based methods, including SVD-LLM \cite{wang2024svd}, Dobi-SVD \cite{wang2025dobi}, Basis Sharing \cite{wang2024basis}, and SAES-SVD \cite{hu2026saes}. We also include earlier methods such as FWSVD \cite{hsu2022language} and ASVD \cite{yuan2023asvd} as additional baselines.

\vspace{-0.05in}
\textbf{Implementation Details.}
The router is implemented as a single linear gating layer. We use AdamW with a learning rate of $2\times10^{-4}$, weight decay of $10^{-3}$, and a cosine learning-rate schedule with 10\% warmup. The maximum sequence length is set to 2048, and training is performed in bfloat16. The global batch size is 64 with gradient accumulation. For router training, we sample 5K prompts from C4 and train the router for 5 epochs, requiring approximately 6 GPU-hours on NVIDIA RTX A6000 GPUs. All latency measurements are conducted on 2 NVIDIA RTX A6000 48GB GPUs.

\begin{table*}[!ht]
\centering
\renewcommand{\arraystretch}{1.08}
\vspace{-0.1in}
\caption{Performance comparison under different compression ratios. Our method is orthogonal to existing SVD-based compression methods and can be integrated into different backbones consistently. Lower is better for Wiki2, PTB, and C4; higher is better for the remaining benchmarks.}
\resizebox{\textwidth}{!}{%
\begin{tabular}{c l c c c c c c c c c c c}
\toprule
\textbf{Compression}
& & \multicolumn{3}{c}{\textbf{PPL} ($\downarrow$)} & \multicolumn{8}{c}{\textbf{Accuracy} ($\uparrow$)} \\
\cmidrule(lr){3-5} \cmidrule(lr){6-13}
\textbf{Ratio} & \multicolumn{1}{c}{\textbf{Method}} &
\textbf{Wiki2} & \textbf{PTB} & \textbf{C4}
& \textbf{Openb.} & \textbf{ARC\_e} & \textbf{ARC\_c} & \textbf{WinoG.} & \textbf{HellaS.} & \textbf{PIQA} & \textbf{MathQA} & \textbf{Avg.} \\
\midrule

0.0 & Baseline
& 5.68 & 8.35 & 7.34
& 0.28 & 0.67 & 0.38 & 0.67 & 0.56 & 0.78 & 0.27 & 0.52 \\
\midrule

\multirow{10}{*}{0.2}
& FWSVD
& 1727 & 2152 & 1511
& 0.09 & 0.11 & 0.06 & 0.05 & 0.08 & 0.10 & 0.05 & 0.08 \\
& ASVD
& 11.14 & 16.55 & 15.93
& 0.25 & 0.53 & 0.27 & 0.64 & 0.41 & 0.68 & 0.24 & 0.43 \\
& SVD-LLM V2
& 7.12 & - & 10.47
& 0.32 & 0.72 & - & 0.70 & 0.52 & 0.75 & 0.24 & - \\
\cline{2-13}
& SVD-LLM
& 7.94 & 16.22 & 15.84
& 0.22 & 0.58 & 0.29 & 0.63 & 0.43 & 0.69 & 0.24 & 0.44 \\
& \cellcolor{gray!15}SVD-LLM + \textbf{PARSE}
& \cellcolor{gray!15}\textbf{7.43} & \cellcolor{gray!15}\textbf{14.37} & \cellcolor{gray!15}\textbf{14.18}
& \cellcolor{gray!15}\textbf{0.26} & \cellcolor{gray!15}\textbf{0.63} & \cellcolor{gray!15}\textbf{0.32} & \cellcolor{gray!15}\textbf{0.66} & \cellcolor{gray!15}\textbf{0.48} & \cellcolor{gray!15}\textbf{0.75} & \cellcolor{gray!15}\textbf{0.25} & \cellcolor{gray!15}\textbf{0.48} \\

& Dobi-SVD
& 8.54 & 14.83 & 10.01
& 0.26 & 0.59 & 0.31 & 0.66 & 0.44 & 0.70 & 0.23 & 0.46 \\
& \cellcolor{gray!15}Dobi-SVD + \textbf{PARSE}
& \cellcolor{gray!15}\textbf{7.79} & \cellcolor{gray!15}\textbf{13.90} & \cellcolor{gray!15}\textbf{9.92}
& \cellcolor{gray!15}\textbf{0.27} & \cellcolor{gray!15}\textbf{0.65} & \cellcolor{gray!15}\textbf{0.35} & \cellcolor{gray!15}\textbf{0.67} & \cellcolor{gray!15}\textbf{0.46} & \cellcolor{gray!15}\textbf{0.76} & \cellcolor{gray!15}\textbf{0.25} & \cellcolor{gray!15}\textbf{0.49} \\

& Basis Sharing
& 7.74 & 17.35 & 15.03
& 0.28 & 0.66 & 0.36 & 0.66 & 0.46 & 0.71 & 0.25 & 0.48 \\
& \cellcolor{gray!15}Basis Sharing + \textbf{PARSE}
& \cellcolor{gray!15}\textbf{7.21} & \cellcolor{gray!15}\textbf{14.98} & \cellcolor{gray!15}\textbf{14.06}
& \cellcolor{gray!15}\textbf{0.28} & \cellcolor{gray!15}\textbf{0.67} & \cellcolor{gray!15}\textbf{0.37} & \cellcolor{gray!15}\textbf{0.67} & \cellcolor{gray!15}\textbf{0.50} & \cellcolor{gray!15}\textbf{0.77} & \cellcolor{gray!15}\textbf{0.27} & \cellcolor{gray!15}\textbf{0.51} \\

& SAES-SVD
& 7.17 & 15.16 & 13.77
& \textbf{0.29} & \textbf{0.68} & 0.36 & 0.65 & 0.45 & 0.75 & 0.25 & 0.49 \\
& \cellcolor{gray!15}SAES-SVD + \textbf{PARSE}
& \cellcolor{gray!15}\textbf{7.01} & \cellcolor{gray!15}\textbf{13.96} & \cellcolor{gray!15}\textbf{12.84}
& \cellcolor{gray!15}0.28 & \cellcolor{gray!15}0.67 & \cellcolor{gray!15}\textbf{0.37} & \cellcolor{gray!15}\textbf{0.67} & \cellcolor{gray!15}\textbf{0.50} & \cellcolor{gray!15}\textbf{0.78} & \cellcolor{gray!15}\textbf{0.27} & \cellcolor{gray!15}\textbf{0.51} \\
\midrule

\multirow{10}{*}{0.4}
& FWSVD
& 18156 & 20990 & 12847
& 0.06 & 0.05 & 0.02 & 0.02 & 0.00 & 0.05 & 0.03 & 0.03 \\
& ASVD
& 1407 & 3292 & 1109
& 0.13 & 0.28 & 0.22 & 0.48 & 0.26 & 0.55 & 0.19 & 0.30 \\
\cline{2-13}
& SVD-LLM
& 13.11 & 63.75 & 49.83
& 0.19 & 0.42 & 0.25 & 0.58 & 0.33 & 0.60 & 0.21 & 0.37 \\
& \cellcolor{gray!15}SVD-LLM + \textbf{PARSE}
& \cellcolor{gray!15}\textbf{12.48} & \cellcolor{gray!15}\textbf{50.17} & \cellcolor{gray!15}\textbf{35.61}
& \cellcolor{gray!15}\textbf{0.22} & \cellcolor{gray!15}\textbf{0.48} & \cellcolor{gray!15}\textbf{0.27} & \cellcolor{gray!15}\textbf{0.62} & \cellcolor{gray!15}\textbf{0.37} & \cellcolor{gray!15}\textbf{0.68} & \cellcolor{gray!15}\textbf{0.24} & \cellcolor{gray!15}\textbf{0.41} \\

& Dobi-SVD
& 13.54 & 46.38 & 23.54
& 0.22 & 0.41 & 0.27 & 0.58 & 0.34 & 0.61 & 0.23 & 0.38 \\
& \cellcolor{gray!15}Dobi-SVD + \textbf{PARSE}
& \cellcolor{gray!15}\textbf{12.33} & \cellcolor{gray!15}\textbf{40.16} & \cellcolor{gray!15}\textbf{20.89}
& \cellcolor{gray!15}\textbf{0.23} & \cellcolor{gray!15}\textbf{0.52} & \cellcolor{gray!15}\textbf{0.29} & \cellcolor{gray!15}\textbf{0.63} & \cellcolor{gray!15}\textbf{0.38} & \cellcolor{gray!15}\textbf{0.69} & \cellcolor{gray!15}\textbf{0.25} & \cellcolor{gray!15}\textbf{0.43} \\

& Basis Sharing
& 12.39 & 55.78 & 41.28
& 0.22 & 0.52 & 0.27 & 0.61 & 0.35 & 0.62 & 0.23 & 0.40 \\
& \cellcolor{gray!15}Basis Sharing + \textbf{PARSE}
& \cellcolor{gray!15}\textbf{10.98} & \cellcolor{gray!15}\textbf{44.49} & \cellcolor{gray!15}\textbf{33.19}
& \cellcolor{gray!15}\textbf{0.24} & \cellcolor{gray!15}\textbf{0.58} & \cellcolor{gray!15}\textbf{0.30} &\cellcolor{gray!15}\textbf{0.65} & \cellcolor{gray!15}\textbf{0.39} & \cellcolor{gray!15}\textbf{0.70} & \cellcolor{gray!15}\textbf{0.25} &  \cellcolor{gray!15}\textbf{0.44} \\

& SAES-SVD
& 10.42 & 45.13 & 32.79
& 0.23 & 0.50 & 0.29 & 0.62 & 0.36 & 0.65 & 0.23 & 0.41 \\
& \cellcolor{gray!15}SAES-SVD + \textbf{PARSE}
& \cellcolor{gray!15}\textbf{9.88} & \cellcolor{gray!15}\textbf{42.13} & \cellcolor{gray!15}\textbf{29.76}
& \cellcolor{gray!15}\textbf{0.25} & \cellcolor{gray!15}\textbf{0.54} & \cellcolor{gray!15}\textbf{0.31} & \cellcolor{gray!15}\textbf{0.66} & \cellcolor{gray!15}\textbf{0.40} & \cellcolor{gray!15}\textbf{0.70} & \cellcolor{gray!15}\textbf{0.25} & \cellcolor{gray!15}\textbf{0.44} \\
\midrule

\multirow{10}{*}{0.6}
& FWSVD
& 32194 & 23575 & 29292
& 0.06 & 0.01 & 0.00 & 0.00 & 0.01 & 0.01 & 0.00 & 0.01 \\
& ASVD
& 57057 & 45218 & 43036
& 0.12 & 0.26 & 0.21 & 0.49 & 0.26 & 0.53 & 0.18 & 0.29 \\
\cline{2-13}
& SVD-LLM
& 53.74 & 438.58 & 345.49
& 0.14 & 0.28 & 0.22 & 0.50 & 0.27 & 0.55 & 0.21 & 0.31 \\
& \cellcolor{gray!15}SVD-LLM + \textbf{PARSE}
& \cellcolor{gray!15}\textbf{27.18} & \cellcolor{gray!15}\textbf{298.46} & \cellcolor{gray!15}\textbf{209.92}
& \cellcolor{gray!15}\textbf{0.18} & \cellcolor{gray!15}\textbf{0.35} & \cellcolor{gray!15}\textbf{0.27} & \cellcolor{gray!15}\textbf{0.55} & \cellcolor{gray!15}\textbf{0.31} & \cellcolor{gray!15}\textbf{0.58} & \cellcolor{gray!15}\textbf{0.24} & \cellcolor{gray!15}\textbf{0.35} \\

& Dobi-SVD
& 46.18 & 238.91 & 190.62
& 0.15 & 0.31 & 0.20 & 0.52 & 0.28 & 0.54 & 0.22 & 0.32 \\
& \cellcolor{gray!15}Dobi-SVD + \textbf{PARSE}
& \cellcolor{gray!15}\textbf{23.28} & \cellcolor{gray!15}\textbf{156.41} & \cellcolor{gray!15}\textbf{126.54}
& \cellcolor{gray!15}\textbf{0.20} & \cellcolor{gray!15}\textbf{0.36} & \cellcolor{gray!15}\textbf{0.29} & \cellcolor{gray!15}\textbf{0.57} & \cellcolor{gray!15}\textbf{0.33} & \cellcolor{gray!15}\textbf{0.60} & \cellcolor{gray!15}\textbf{0.24} & \cellcolor{gray!15}\textbf{0.37} \\

& Basis Sharing
& 43.81 & 352.64 & 250.19
& 0.15 & 0.34 & 0.21 & 0.53 & 0.29 & 0.54 & 0.21 & - \\
& \cellcolor{gray!15}Basis Sharing + \textbf{PARSE}
& \cellcolor{gray!15}\textbf{21.56} & \cellcolor{gray!15}\textbf{175.64} & \cellcolor{gray!15}\textbf{101.21}
& \cellcolor{gray!15}\textbf{0.21} & \cellcolor{gray!15}\textbf{0.40} & \cellcolor{gray!15}\textbf{0.30} & \cellcolor{gray!15}\textbf{0.64} & \cellcolor{gray!15}\textbf{0.37} & \cellcolor{gray!15}\textbf{0.65} & \cellcolor{gray!15}\textbf{0.24} & \cellcolor{gray!15}\textbf{0.40} \\

& SAES-SVD
& 22.01 & 116.83 & 93.97
& 0.16 & 0.33 & 0.25 & 0.52 & 0.30 & 0.54 & 0.23 & 0.34 \\
& \cellcolor{gray!15}SAES-SVD + \textbf{PARSE}
& \cellcolor{gray!15}\textbf{19.83} & \cellcolor{gray!15}\textbf{96.17} & \cellcolor{gray!15}\textbf{81.46}
& \cellcolor{gray!15}\textbf{0.22} & \cellcolor{gray!15}\textbf{0.40} & \cellcolor{gray!15}\textbf{0.31} & \cellcolor{gray!15}\textbf{0.63} & \cellcolor{gray!15}\textbf{0.39} & \cellcolor{gray!15}\textbf{0.65} & \cellcolor{gray!15}\textbf{0.24} & \cellcolor{gray!15}\textbf{0.44} \\
\bottomrule
\end{tabular}%
}
\label{tab:main_table}
\vspace{-0.2in}
\end{table*}

\vspace{-0.1in}
\subsection{Main Results}
\label{sec:main_results}
\textbf{Accuracy.}
Table~\ref{tab:main_table} reports PPL and zero-shot accuracy results on LLaMA-7B across multiple SVD-based compression backbones and compression ratios. Overall, PARSE consistently achieves lower PPL and higher average zero-shot accuracy than the corresponding baselines. 
At compression ratio 0.2, PARSE improves the average accuracy of SVD-LLM, Dobi-SVD, Basis Sharing, and SAES-SVD by 4.0, 3.0, 3.0, and 2.0 percentage points, respectively. The improvements become larger under more aggressive compression. For example, at compression ratio 0.6, PARSE improves SAES-SVD from 0.34 to 0.44 average accuracy. Similar gains are also observed in language modeling performance, where PARSE reduces C4 PPL from 250.19 to 101.21 on Basis Sharing and from 93.97 to 81.46 on SAES-SVD. 
These results demonstrate that prompt-aware rank selection effectively preserves model quality under aggressive SVD compression.

\vspace{-0.1in}
\begin{figure*}[h]
  \centering
  \includegraphics[width=0.93\linewidth]
  {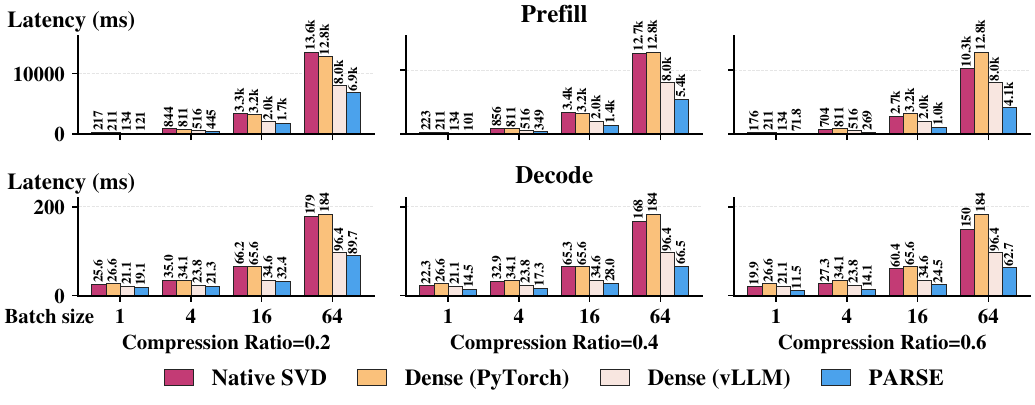}
  \vspace{-0.15in}
  \caption{Prefill latency (ms) and Decode latency (ms) of each token of Native SVD, Dense(Pytorch), Dense(vLLM) and PARSE on LLaMA-7B, across compression ratios and batch sizes.}
  \label{fig:main_latency_ratio_grid}
  \vspace{-0.1in}
\end{figure*}

\begin{wraptable}[8]{r}{0.55\textwidth}
\vspace{-0.3in}
\centering
\caption{PPL and average accuracy comparison of SAES-SVD and PARSE at compression ratio 0.2.}
\resizebox{0.55\textwidth}{!}{
\begin{tabular}{l l c c c c}
\toprule
& & \multicolumn{3}{c}{\textbf{PPL} ($\downarrow$)} & \textbf{Avg. Accuracy} ($\uparrow$) \\
\cmidrule(lr){3-5} \cmidrule(lr){6-6}
\textbf{Model} & \textbf{Method} 
& \textbf{WikiText-2} & \textbf{PTB} & \textbf{C4} & \textbf{Zero-shot Tasks} \\
\midrule

\multirow{2}{*}{LLaMA-13B}
& SAES-SVD
& 6.34 & 35.76 & 11.47 & 0.51 \\
& \cellcolor{gray!15}SAES-SVD + \textbf{PARSE}
& \cellcolor{gray!15}\textbf{6.05} 
& \cellcolor{gray!15}\textbf{30.27} 
& \cellcolor{gray!15}\textbf{9.31} 
& \cellcolor{gray!15}\textbf{0.55} \\

\midrule

\multirow{2}{*}{LLaMA-30B}
& SAES-SVD
& 5.49 & 29.78 & 9.16 & 0.57 \\
& \cellcolor{gray!15}SAES-SVD + \textbf{PARSE}
& \cellcolor{gray!15}\textbf{5.21} 
& \cellcolor{gray!15}\textbf{27.04} 
& \cellcolor{gray!15}\textbf{8.27} 
& \cellcolor{gray!15}\textbf{0.59} \\

\midrule

\multirow{2}{*}{Qwen2.5-7B}
& SAES-SVD
& 8.23 & 17.64 & 19.47 & 0.58 \\
& \cellcolor{gray!15}SAES-SVD + \textbf{PARSE}
& \cellcolor{gray!15}\textbf{7.83} 
& \cellcolor{gray!15}\textbf{15.52} 
& \cellcolor{gray!15}\textbf{16.57} 
& \cellcolor{gray!15}\textbf{0.61} \\

\bottomrule
\end{tabular}
}
\label{tab:cross_model_validation}
\vspace{-0.15in}
\end{wraptable}

\textbf{Performance on Larger and Newer Models.}
Table~\ref{tab:cross_model_validation} 
compares SAES-SVD with PARSE
under compression ratio 0.2 on larger models (i.e., LLaMA-13B and LLaMA-30B) and Qwen2.5-7B model. PARSE consistently improves both PPL and zero-shot accuracy across different models. 
For example, on LLaMA-13B, PARSE reduces WikiText-2 PPL from 6.34 to 6.05 and improves average accuracy from 0.51 to 0.55. Similar gains are observed on LLaMA-30B and Qwen2.5-7B, where PARSE further reduces PTB PPL from 29.78 to 27.04 and C4 PPL from 19.47 to 16.57, respectively. 
These results demonstrate that PARSE generalizes effectively across both large-scale LLaMA models and newer models.

\textbf{Latency.}
Figure~\ref{fig:main_latency_ratio_grid} compares the prefill and decode latency of Native SVD, Dense (PyTorch), Dense (vLLM), and PARSE on LLaMA-7B.
All SVD-based baselines in Table~\ref{tab:main_table} share the same two-MatMul execution path, so we report them as a single Native SVD baseline.
Overall, PARSE consistently achieves lower latency across compression ratios and batch sizes.
The advantage becomes more substantial at larger batch sizes.
For prefilling, PARSE achieves up to $2.5\times$ speedup over Native SVD and reaches 4.1s latency at batch size 64 under compression ratio 0.6.
For decoding at same settings, PARSE achieves up to $2.4\times$ speedup and reduces per-token latency to 62.7ms under the same setting.
PARSE also outperforms Dense (vLLM) by up to $2.0\times$ in prefilling under the strongest compression setting.
These results demonstrate that PARSE provides practical inference speedup while preserving the quality gains of prompt-aware rank selection.

\vspace{-0.1in}
\subsection{Ablation Studies}
\vspace{-0.1in}

\textbf{Effects of Rank Retrieval and Reuse.}
Table~\ref{tab:ablation_inference_opt} evaluates the effect of the proposed rank retrieval and reuse strategies on model inference. The router-based PARSE achieves the best overall perplexity and accuracy but introduces substantial overhead from dynamic routing at every decoding step. 
On the SVD-LLM backbone, PARSE (Router) improves WikiText-2 perplexity from 7.94 to 7.16 and lifts average zero-shot accuracy from 0.44 to 0.52, but prefill latency rises from 217.08\,ms to 516.49\,ms. Rank retrieval reduces prefill latency to 121.64\,ms while preserving competitive perplexity and accuracy, and rank reuse cuts decode latency from 68.14\,ms to 20.78\,ms by sharing routing decisions across generated tokens. Combining both yields PARSE (Full), which achieves the lowest overall latency on both SVD-LLM (19.06\,ms) and SAES-SVD (18.94\,ms). Despite a slight quality drop relative to the router-only variant, PARSE (Full) still substantially outperforms the original compressed baselines in both ppl and reasoning accuracy, delivering up to 1.8$\times$ prefill and 1.4$\times$ decode speedup when batch size is 1.

\begin{figure}[t]
  \centering
  \includegraphics[width=0.93\linewidth]{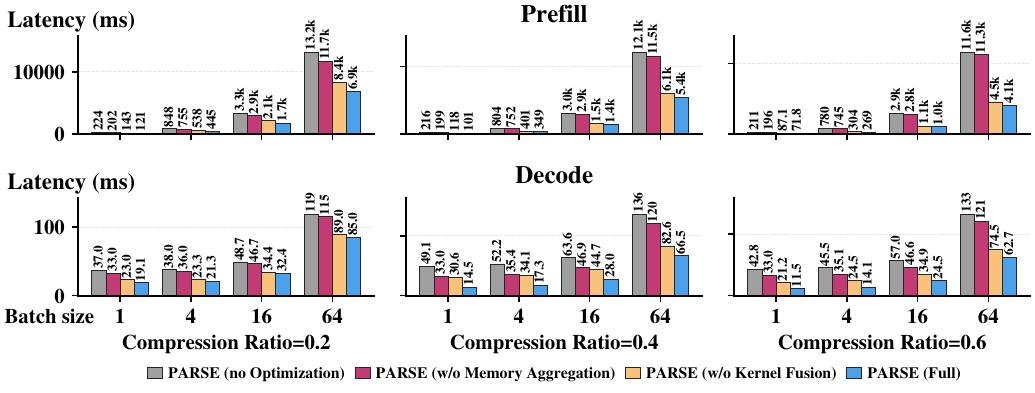}
  \vspace{-0.15in}
  \caption{Prefill latency (ms) and decode latency (ms) of each token of \textbf{PARSE} under different hardware optimization configurations on LLaMA-7B.
  }
  \vspace{-0.2in}
  \label{fig:system_efficiency}
\end{figure}

\begin{wraptable}[11]{r}{0.55\textwidth}
\centering
\vspace{-0.25in}
\caption{Effects of inference optimization components on LLaMA-7B with compression ratio 0.2. Batch size is 1.}
\vspace{0.05in}
\resizebox{0.54\textwidth}{!}{
\begin{tabular}{l c c c c c c}
\toprule
& \multicolumn{3}{c}{\textbf{PPL} ($\downarrow$)} & \textbf{Avg. Accuracy} ($\uparrow$) & \multicolumn{2}{c}{\textbf{Latency (ms)} $\downarrow$} \\
\cmidrule(lr){2-4} \cmidrule(lr){5-5} \cmidrule(lr){6-7}
\textbf{Method} & \textbf{Wikitext-2} & \textbf{PTB} & \textbf{C4} & \textbf{Zero-shot Tasks} & \textbf{Prefill} & \textbf{Decode} \\
\midrule
SVD-LLM                                            & 7.94 & 16.22 & 15.84 & 0.44 & 217.08 & 25.56 \\
\quad + \textbf{PARSE} (Router)                        & \textbf{7.16} & \textbf{13.75} & \textbf{12.41} & \textbf{0.52} & 516.49 & 68.14 \\
\quad\quad + Rank Retrieval                        & 7.43 & 14.37 & 14.18 & 0.50 & 121.64 & 68.19 \\
\quad\quad + Rank Reuse                            & \textbf{7.16} & \textbf{13.75} & \textbf{12.41} & 0.49 & 517.61 & 20.78 \\
\rowcolor{gray!15}
\quad + \textbf{PARSE} (Full)                               & 7.43 & 14.37 & 14.18 & 0.48 & \textbf{120.72} & \textbf{19.06} \\
\midrule
SAES-SVD                                           & 7.17 & 15.16 & 13.77 & 0.49 & 216.69 & 26.18 \\
\quad + \textbf{PARSE} (Router)                        & \textbf{6.82} & \textbf{12.84} & \textbf{11.96} & \textbf{0.53} & 517.56 & 70.15 \\
\quad\quad + Rank Retrieval                        & 7.01 & 13.96 & 12.84 & \textbf{0.53} & 123.84 & 71.91 \\
\quad\quad + Rank Reuse                            & \textbf{6.82} & \textbf{12.84} & \textbf{11.96} & 0.52 & 516.28 & 19.21 \\
\rowcolor{gray!15}
\quad + \textbf{PARSE} (Full)                               & 7.01 & 13.96 & 12.84 & 0.51 & \textbf{121.11} & \textbf{18.94} \\
\bottomrule
\end{tabular}
}
\label{tab:ablation_inference_opt}
\end{wraptable}

\textbf{Effect of Memory Aggregation and Kernel Fusion.}
Figure~\ref{fig:system_efficiency} ablates the two system optimizations used by PARSE. 
Without optimization, prompt-aware rank selection introduces substantial runtime overhead from scattered expert accesses and repeated SVD factorization kernels, reaching 13.2s prefill latency and 119ms decode latency at batch size 64 under compression ratio 0.2. 
Memory aggregation improves efficiency by placing selected rank experts into contiguous memory regions, reducing batch-64 prefill latency from 13.2s to 8.4s under compression ratio 0.2. Kernel fusion further minimizes redundant kernel launches across factorized projections. Combined together, they further reduce prefill latency to 6.9s and decode latency to 85.0ms, consistently achieving the lowest latency across all compression ratios and batch sizes.


\begin{wraptable}[7]{r}{0.5\textwidth}
\vspace{-0.25in}
\centering
\caption{Impact of router training data on LLaMA-7B with SAES-SVD at compression ratio 0.2.}
\resizebox{0.5\textwidth}{!}{
\begin{tabular}{l c c c c}
\toprule
& \multicolumn{3}{c}{\textbf{PPL} ($\downarrow$)} & \textbf{Avg. Accuracy} ($\uparrow$) \\
\cmidrule(lr){2-4} \cmidrule(lr){5-5}
\textbf{Train. Dataset} & \textbf{WikiText-2} & \textbf{PTB} & \textbf{C4} & \textbf{Zero-shot Tasks} \\
\midrule
WikiText-2        & 7.05 & 14.09 & 13.21 & 0.50 \\
PTB               & 7.12 & 14.97 & 13.87 & 0.48 \\
\rowcolor{gray!15}
\textbf{C4} & \textbf{7.01} & \textbf{13.96} & \textbf{12.84} & \textbf{0.51} \\
\bottomrule
\end{tabular}
}
\label{tab:ablation_router_data}
\end{wraptable}

\textbf{Impact of Router Training Data.}
Table~\ref{tab:ablation_router_data} evaluates how the router training corpus affects prompt-aware rank selection. 
Training the router on C4 achieves the best overall performance, with the lowest perplexity on WikiText-2, PTB, and C4, and the highest average zero-shot accuracy of 0.51. 
In contrast, training on WikiText-2 slightly degrades PTB and C4 perplexity, while training on PTB causes a larger drop, increasing perplexity from 13.96 to 14.97 on PTB and from 12.84 to 13.87 on C4, with average accuracy falling from 0.51 to 0.48. 
This shows that the router can inherit the domain bias of its training corpus, even though the SVD factors are fixed. 
A diverse corpus such as C4 exposes the router to broader activation patterns, allowing it to learn rank selection rules that transfer more reliably across downstream tasks.

\vspace{-0.1in}

\section{Conclusion}

\vspace{-0.1in}

We present PARSE, a prompt-aware rank selection framework for SVD-based LLM compression. Unlike static rank truncation, PARSE reformulates singular components as independent rank experts and learns prompt-aware routing to dynamically select rank components for each input. To reduce dependence on calibration information, the router is trained against dense-model outputs on a large-scale corpus, enabling robust rank selection across diverse downstream tasks. We further introduce rank retrieval and reuse strategies to eliminate routing overhead at inference, together with memory aggregation and kernel fusion for efficient deployment. Extensive experiments across multiple SVD-based baselines, model scales, and reasoning benchmarks show that PARSE consistently improves both perplexity and zero-shot accuracy while reducing inference latency.


\clearpage

\newpage

\bibliographystyle{unsrt}

\bibliography{ref}

\newpage

\appendix

\section{Related Works}
\paragraph{Large Language Model Compression.}
Large language model compression has been widely studied to reduce memory and inference cost while preserving performance~\cite{zhu2024survey,xu2023survey}.
Knowledge distillation~\cite{hinton2015distilling,gu2023minillm} compresses models by transferring knowledge from a large teacher to a compact student, but it requires an additional training stage and exhibits limited generalization capability~\cite{xu2024survey}.
Pruning avoids this training cost by directly modifying the weight structure of a pretrained model.
Unstructured pruning~\cite{frantar2023sparsegpt,sun2023simple} achieves high sparsity but yields irregular weight patterns that limit practical acceleration on standard hardware~\cite{jeong2025enabling,dave2021hardware}.
Structured pruning~\cite{ma2023llm,ashkboos2024slicegpt} addresses this by removing entire channels or attention heads to ensure deployment efficiency, yet it faces a harder performance-efficiency trade-off~\cite{guo2025slimllm}.
Quantization~\cite{frantar2022gptq,lin2024awq,xiao2023smoothquant,yao2022zeroquant} eases this trade-off by compressing the numerical precision of weights rather than their structure, but its effectiveness depends on hardware support and degrades under aggressive low-bit settings~\cite{li2025quantization}.
Low-rank approximation~\cite{yuan2023asvd,wang2024svd} sidesteps these hardware dependencies by preserving regular dense linear operations, while also avoiding large-scale fine-tuning as a post-training compression strategy~\cite{zhu2024survey}, yet existing methods produce static compressed models that do not adapt to different inputs.

\paragraph{SVD-based Compression for LLMs.}
SVD-based compression has emerged as an important direction for reducing the parameter count and inference cost of LLMs.
Standard SVD minimizes truncation error in weight space, while inference quality depends on activation reconstruction rather than weights~\cite{wang2024svd}.
To bridge this gap, FWSVD~\cite{hsu2022language} uses Fisher information to weight the importance of individual parameters, ASVD~\cite{yuan2023asvd} absorbs activation outliers into weight matrices via activation statistics before decomposition, and SVD-LLM~\cite{wang2024svd} applies data whitening to align singular values more directly with activation reconstruction loss.
Beyond optimizing the decomposition process, several works focus on refining the truncation.
Along this direction, AdaSVD~\cite{li2025adasvd} alternately updates singular matrices to compensate for truncation error, DipSVD~\cite{ding2025dipsvd} jointly considers local and global importance to better identify components to retain, and Dobi-SVD~\cite{wang2025dobi} makes the truncation process differentiable to enable more principled rank selection.
While the above methods overlook inter-layer heterogeneity, another line of work takes a cross-layer perspective.
Basis Sharing~\cite{wang2024basis} exploits cross-layer redundancy by representing weight matrices as linear combinations of shared basis vectors with layer-specific coefficients.
D-Rank~\cite{mi2025layer} measures per-layer information density via effective rank and allocates rank budgets under a static global compression ratio.
SAES-SVD~\cite{hu2026saes} further addresses the accumulated error that propagates across layers by jointly suppressing both per-layer reconstruction error and cross-layer error during compression.

Although these works perform rank truncation for model compression, the truncation is typically determined during calibration, resulting in a static rank selection. This static rank is then uniformly applied to all prompts during inference, without accounting for variability across different input prompts. Motivated by this bottleneck, as well as by the router selection mechanism of Mixture-of-Experts (MoE) models~\cite{lepikhin2020gshard,fedus2022switch,cai2024survey}, we propose to dynamically select ranks via a router based on the input prompt to achieve better performance for different inputs, while maintaining the same compression ratio as baseline methods. This routing paradigm has been widely adopted in modern LLM architectures~\cite{jiang2024mixtral,liu2024deepseek} and has even been retrofitted into dense models without pretraining from scratch~\cite{zhang2022moefication}.

\section{Limitations}
\label{sec:limitations}
PARSE is evaluated on open-source decoder-only LLMs and standard language modeling and zero-shot reasoning benchmarks; its behavior on multimodal models, encoder-decoder models, and instruction-tuned production systems remains to be further studied. The current implementation trains routers for SVD-compressed weights and relies on prompt-level rank patterns, so the benefit may depend on the stability of rank selection across prompts and decoding steps. In addition, although rank retrieval and reuse reduce online router overhead, constructing and storing the rank-pattern cache introduces extra offline preprocessing and memory cost. Finally, our latency measurements are conducted on NVIDIA RTX A6000 GPUs, and the absolute speedups may vary across hardware backends and kernel implementations.

\newpage

\section{More Ablation Studies}
\label{sec:more_ablation}

\begin{wraptable}[8]{r}{0.7\textwidth}
\vspace{-0.25in}
\centering
\caption{Impact of different router designs on LLaMA-7B by SAES-SVD at compression ratio 0.2.}
\resizebox{0.69\textwidth}{!}{
\begin{tabular}{l c c c c c}
\toprule
& & \multicolumn{3}{c}{\textbf{PPL} ($\downarrow$)} & \textbf{Avg. Accuracy} ($\uparrow$) \\
\cmidrule(lr){3-5} \cmidrule(lr){6-6}
\textbf{Router Architecture} & \textbf{Params} ($\downarrow$) & \textbf{WikiText-2} & \textbf{PTB} & \textbf{C4} & \textbf{Zero-shot Tasks} \\
\midrule
MLP (2-layer, ReLU)      & 8423.21M & 7.03 & 13.99 & \textbf{12.83} & 0.50 \\
MLP (2-layer, GELU)      & 8423.21M & 7.01 & 14.08 & 12.84 & 0.50 \\
Attention-block          & 30721.41M & 7.11 & 14.35 & 13.11 & 0.47 \\
\rowcolor{gray!15}
\textbf{Single Linear Layer}   & \textbf{4664.07M} & \textbf{7.01} & \textbf{13.96} & 12.84 & \textbf{0.51} \\
\bottomrule
\end{tabular}
}
\label{tab:ablation_router}
\vspace{-0.2in}
\end{wraptable}

\textbf{Impact of Router Architecture.}
Table~\ref{tab:ablation_router} compares different router architectures on SAES-SVD with compression ratio 0.2. 
The single linear router achieves the best overall trade-off, using 4.66B router parameters while obtaining the lowest PTB perplexity of 13.96 and the highest average zero-shot accuracy of 0.51. 
Increasing router capacity does not lead to consistent gains: two-layer MLP routers nearly double the parameter count to 8.42B but provide no improvement in average accuracy, and the attention-block router further increases the parameter count to 30.72B while degrading perplexity on all three datasets and reducing average accuracy to 0.47. 
These results show that prompt-aware rank selection does not require a high-capacity router. 
A single linear projection is sufficient to separate useful rank components from the input activation, while larger routers introduce unnecessary parameters and become harder to optimize under the same post-training budget.

\begin{wraptable}[5]{r}{0.6\textwidth}
\vspace{-0.3in}
\centering
\caption{Ablation on vanilla SVD without orthogonal baselines on LLaMA-7B at compression ratio 0.2.}
\resizebox{0.6\textwidth}{!}{
\begin{tabular}{lcccc}
\toprule
\textbf{Method} & \textbf{WikiText-2} $\downarrow$ & \textbf{PTB} $\downarrow$ & \textbf{C4} $\downarrow$ & \textbf{Avg. Acc.} $\uparrow$ \\
\midrule
Vanilla SVD          & 20222.35 & 20749.99 & 19014.54 & 0.30 \\
Vanilla SVD + \textbf{PARSE} & \textbf{301.27} & \textbf{512.42} & \textbf{350.26} & \textbf{0.33} \\
\bottomrule
\end{tabular}
}
\label{tab:ablation_decomp}
\end{wraptable}

\textbf{Effectiveness without Orthogonal SVD Enhancements.}
Table~\ref{tab:ablation_decomp} isolates the effect of PARSE by applying it directly to vanilla SVD, without whitening, error compensation, basis sharing, or any decomposition-level enhancement. 
Vanilla SVD collapses under aggressive compression, reaching perplexities above $1.9\times 10^4$ on all three language modeling benchmarks. 
However, after incorporating PARSE, the perplexity is dramatically improved. For example, on the WikiText-2 dataset, vanilla SVD combined with PARSE achieves a perplexity of only 301.27, reducing the perplexity of vanilla SVD by more than 19,000. This demonstrates that PARSE itself can achieve strong compression performance without relying on sophisticated SVD compression baselines, while also highlighting the necessity of dynamically selecting prompt-aware ranks during inference.

\begin{wraptable}[8]{r}{0.5\textwidth}
\vspace{-0.3in}
\centering
\caption{Impact of SVD calibration data on LLaMA-7B with SAES-SVD at compression ratio 0.2.}
\vspace{0.05in}
\resizebox{0.5\textwidth}{!}{
\begin{tabular}{l c c c c}
\toprule
& \multicolumn{3}{c}{\textbf{PPL} ($\downarrow$)} & \textbf{Avg. Accuracy} ($\uparrow$) \\
\cmidrule(lr){2-4} \cmidrule(lr){5-5}
\textbf{Calib. Dataset} & \textbf{WikiText-2} & \textbf{PTB} & \textbf{C4} & \textbf{Zero-shot Tasks} \\
\midrule
WikiText-2        & \textbf{6.98} & 14.07 & 13.04 & 0.51 \\
PTB               & 7.03 & 14.09 & 13.11 & 0.50 \\
C4                & 7.01 & \textbf{13.96} & \textbf{12.84} & \textbf{0.51} \\
MathQA            & 7.05 & 14.10 & 13.13 & 0.50 \\
\bottomrule
\end{tabular}
}
\label{tab:ablation_training_data}
\end{wraptable}

\textbf{Robustness to SVD Calibration Data.}
Table~\ref{tab:ablation_training_data} evaluates whether PARSE remains sensitive to the calibration dataset used by the underlying SVD decomposition. 
We vary the SVD calibration data among WikiText-2, PTB, C4, and MathQA, while keeping the router training corpus fixed to C4.
We include PTB and MathQA as stress-test calibration sources because they produce the largest calibration-induced degradation in Observation~2.
PARSE maintains stable performance across all calibration choices: WikiText-2 perplexity stays within 6.98--7.05, PTB within 13.96--14.10, C4 within 12.84--13.13, and average zero-shot accuracy within 0.50--0.51. 
This contrasts with static SVD truncation, where the calibration set directly determines which rank components are retained and therefore causes cross-domain degradation. 
By training the router against dense outputs on a diverse corpus, PARSE reduces the dependence on the calibration information encoded in the SVD factors and selects rank components according to the input prompt instead.

\begin{wrapfigure}[11]{r}{0.65\textwidth}
    \centering
    \vspace{-0.25in}
    \includegraphics[width=0.65\textwidth]{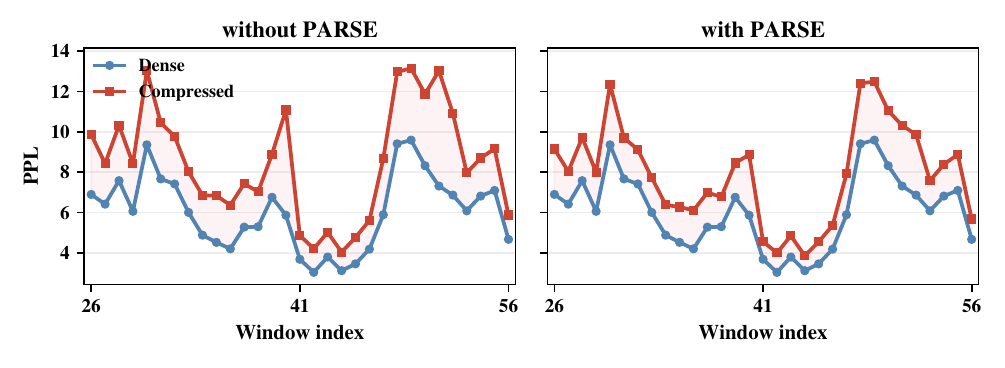}
    \vspace{-0.35in}
    \caption{
    Per-window perplexity on WikiText-2 for the dense and compressed LLaMA-7B by SVD-LLM without PARSE (left) and with PARSE (right). 
    }
    \label{fig:ppl_spike_ablation}
    \vspace{-0.35in}
\end{wrapfigure}

\textbf{Prompt PPL Stability Analysis.}
Figure~\ref{fig:ppl_spike_ablation} extends the per-prompt PPL analysis from 
Section~\ref{sec:obs} by comparing the compressed model with and without PARSE. 
Without PARSE, the compressed model exhibits pronounced perplexity spikes at 
certain windows that far exceed the dense model, consistent with 
Figure~\ref{fig:ppl_spike}. With PARSE, these spikes are substantially 
reduced and the compressed model tracks the dense model much more closely 
across windows, confirming that prompt-aware rank selection mitigates the 
localized degradation caused by static rank truncation.

\section{Existing Assets and Licenses}
\label{sec:assets_licenses}

We use publicly available models, datasets, and evaluation tools only for research evaluation. 
Table~\ref{tab:assets_licenses} summarizes the external assets used in this work and the corresponding license or access terms reported by the original providers. 
We cite the original papers or repositories in the main text and follow the license, access, and acceptable-use terms specified by each asset provider.

\begin{table}[h]
\centering
\caption{External assets used in this work. We use these assets only for research evaluation and follow the license or access terms specified by the original providers.}
\label{tab:assets_licenses}
\small
\setlength{\tabcolsep}{4pt}
\renewcommand{\arraystretch}{1.12}
\begin{tabularx}{\linewidth}{p{0.22\linewidth} p{0.30\linewidth} X}
\toprule
\textbf{Asset} & \textbf{Usage in this paper} & \textbf{License / access terms} \\
\midrule

LLaMA family~\cite{touvron2023llama}
& Target LLMs for SVD compression and evaluation
& Meta LLaMA license and acceptable-use terms \\

Qwen2.5~\cite{hui2024qwen2}
& Target LLM for SVD compression and evaluation
& Apache License 2.0 \\

WikiText-2~\cite{merity2016pointer}
& Language modeling evaluation
& Creative Commons Attribution-ShareAlike \\

Penn Treebank (PTB)~\cite{marcus1993building}
& Language modeling evaluation
& Linguistic Data Consortium license / access terms \\

C4~\cite{raffel2020exploring}
& Router training and language modeling evaluation
& ODC-BY; also subject to Common Crawl terms of use \\

OpenBookQA~\cite{mihaylov2018can}
& Zero-shot reasoning evaluation
& Apache License 2.0 \\

AI2 ARC~\cite{clark2018think}
& Zero-shot reasoning evaluation
& Creative Commons Attribution-ShareAlike \\

WinoGrande~\cite{sakaguchi2021winogrande}
& Zero-shot reasoning evaluation
& CC-BY for the dataset; Apache License 2.0 for the codebase \\

HellaSwag~\cite{zellers2019hellaswag}
& Zero-shot reasoning evaluation
& MIT License \\

PIQA~\cite{bisk2020piqa}
& Zero-shot reasoning evaluation
& License metadata not specified by the dataset card; used for research evaluation with citation \\

MathQA~\cite{amini2019mathqa}
& Zero-shot reasoning evaluation
& Apache License 2.0 \\

LM-Evaluation-Harness~\cite{eval-harness}
& Zero-shot evaluation framework
& MIT License \\

\bottomrule
\end{tabularx}
\vspace{-0.1in}
\end{table}


\newpage

\end{document}